\documentclass[letterpaper, 10 pt, conference]{ieeeconf}  % Comment this line out if you need a4paper

\IEEEoverridecommandlockouts                              % This command is only needed if 
\usepackage{graphicx} % for pdf, bitmapped graphics files
\usepackage{multirow}
\usepackage{booktabs}
\usepackage{pifont}
\usepackage{amsmath} % assumes amsmath package installed
\usepackage{amssymb}  % assumes amsmath package installed
\usepackage{mathtools}
\usepackage{cite}

\usepackage{hyperref}

\usepackage{color}
\usepackage{xcolor}

\usepackage[abs]{overpic}
\usepackage{pict2e}
\usepackage[pdf]{graphviz}

\newcommand{\R}{\mathbb{R}}
\newcommand{\N}{\mathcal{N}}
\newcommand{\p}{\text{p}}
\newcommand{\SOT}{\text{SO(3)}}
\newcommand{\SET}{\text{SE(3)}}
\newcommand{\tip}{\text{tip}}
\newcommand{\tcp}{\text{tcp}}
\newcommand{\base}{\text{base}}

\DeclareMathOperator*{\argmin}{arg\,min}

\title{\LARGE \bf%
Fixture calibration with guaranteed bounds\\% 
from a few correspondence-free surface points
}

\author{Rasmus Laurvig Haugaard, Yitaek Kim and Thorbjørn Mosekjær Iversen% <-this % stops a space
\thanks{
All authors are from SDU Robotics, Maersk Mc-Kinney Moller Institute, University of Southern Denmark.
The authors gratefully acknowledge the support from Innovation Fund Denmark through the project MADE Fast.
\newline
{\tt\small \{rlha,yik,thmi\}@mmmi.sdu.dk}}% <-this % stops a space
}

\begin{document}

\maketitle
% plain: numbers, empty: no numbers
% should by empty for camera ready, plain for arxiv
\thispagestyle{plain}
\pagestyle{plain}

\begin{abstract}
Calibration of fixtures in robotic work cells is essential but also time consuming and error-prone, and poor calibration can easily lead to wasted debugging time in downstream tasks. Contact-based calibration methods let the user measure points on the fixture's surface with a tool tip attached to the robot's end effector. Most such methods require the user to manually annotate correspondences on the CAD model, however, this is error-prone and a cumbersome user experience. We propose a correspondence-free alternative: The user simply measures a few points from the fixture's surface, and our method provides a tight superset of the poses which could explain the measured points. This naturally detects ambiguities related to symmetry and uninformative points and conveys this uncertainty to the user. Perhaps more importantly, it provides \textit{guaranteed} bounds on the pose. The computation of such bounds is made tractable by the use of a hierarchical grid on SE(3). Our method is evaluated both in simulation and on a real collaborative robot, showing great potential for easier and less error-prone fixture calibration.
\textcolor{blue}{\textit{\href{https://sites.google.com/view/ttpose}{sites.google.com/view/ttpose}}}
\end{abstract}
\section{Introduction}
The use of robots in industrial production promises high flexibility, 
however in practice, this flexibility is realized through reconfiguration of robotic work cells, 
which in many cases requires expert knowledge. 
For small and medium enterprises, which often lack in-house robotics experts, the promised flexibility can only be achieved if the reconfiguration process is fast and easily done by non-experts.

\begin{figure}[t]
    \centering
    \includegraphics[width=\linewidth,trim={0 0 0 0.1cm},clip]{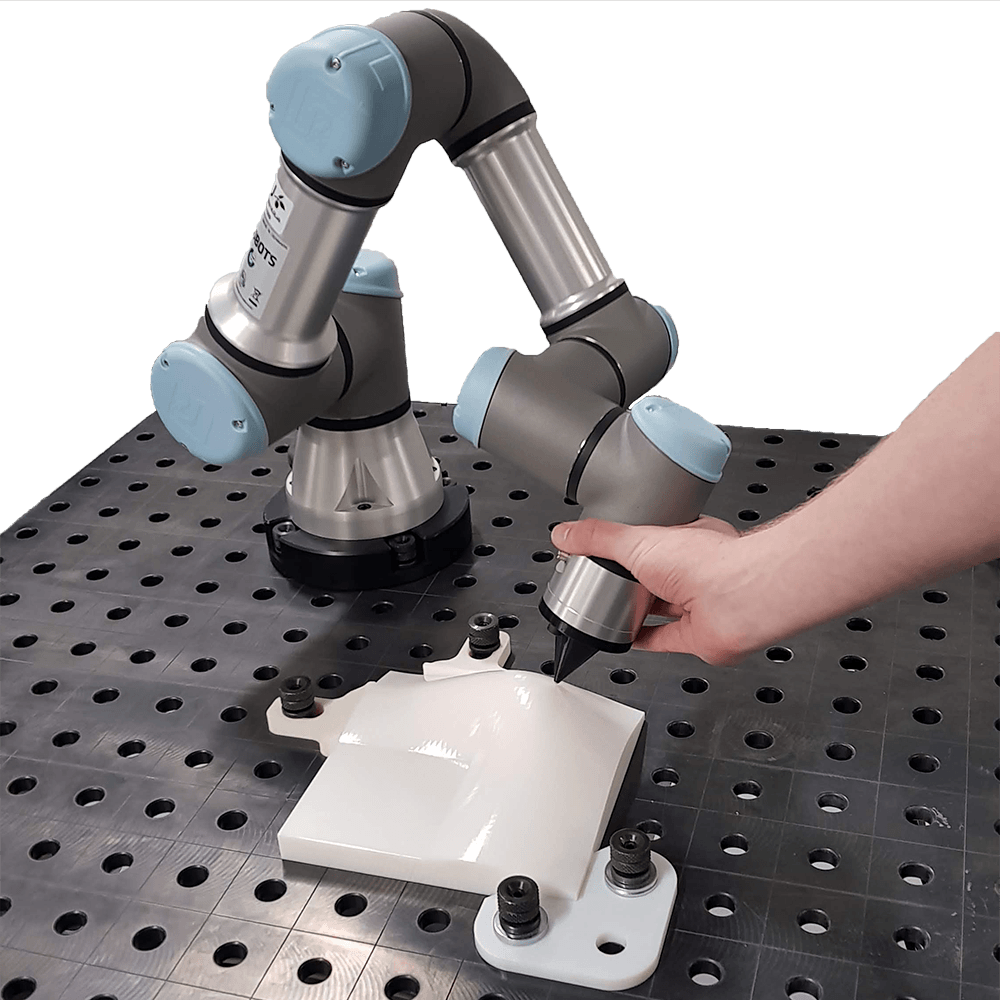}
    \caption{
        A user moves a robot to measure a few points on a fixture using a calibrated tool-tip attached to the end-effector of the robot.
        The tool consist of a 3D-printed mount with a 3 mm steel ball at the tip.
        The fixture is a 3D-printed MATLAB logo, fixated to the table using additional fixtures.
    }
    \label{fig:system}
\end{figure}

An essential part of the configuration process is the calibration of the positions and orientations, also referred to as poses, of the fixtures in the work cell.
Calibrating fixture poses can enable transfer of fixture-related skills which have either been programmed or demonstrated in another configuration.
Accurate calibration with consumer measuring devices, like rulers, is near-infeasible, 
and vision based solutions require additional expensive equipment and setup. 
A promising alternative is contact-based pose estimation, in which the user moves a tool tip, which is attached to the robot's end effector, to various points on the surface of a fixture.
The surface point coordinates are measured in the robot's frame of reference using the robot's forward kinematics and the sampled points are then used to infer the fixture's pose relative to the robot.

Existing contact-based methods rely on correspondences to infer the pose.
Most methods require the user to manually annotate these correspondences, either in form of point-to-point, point-to-line and/or point-to-plane correspondences.
Repeatedly going back and forth between demonstrating points with the robot and annotating correspondences in a computer program is not only a cumbersome user experience; It also requires the user to identify features on the fixture which the user can measure with low uncertainty. %- and for objects with smoothly varying surfaces, such points are simply not available.
This process results in uncertainties that are primarily related to human ability, and it is thus difficult to assume a maximum correspondence error, especially for non-expert users.

In this work, we greatly simplify the user experience by not requiring correspondences.
The user is only required to measure a few correspondence-free points on the object surface. %, which can be done on any object, regardless of the object features.
The user thus \textit{only} has to make sure, that the measured points are in fact from the fixture's surface and not, for example, from another object; so it is reasonable to assume this to hold, even for non-expert users.
The measurement uncertainties are thus restricted to forward kinematics and tool tip calibration, and a conservative estimate of the maximum error can be determined experimentally.

%While accurate calibrations are achievable for feature rich fixtures when performed by an expert, it is often unclear how accurate a calibration is when performed by a non-expert for objects or fixtures where point correspondences are unclear or ambiguous due to a lack of well-defined surface features.

%Our method tackles this problem by treating the fixture calibration as a pose estimation problem with unknown correspondences. 

Given a CAD model and surface measurements with bounded errors, our method finds a tight superset of the set of poses which could explain the measured surface points.
The superset is \textit{guaranteed} to include the true pose, 
and we can thus infer bounds on positional and rotational error.

Our method consists of a search through a hierarchical grid on the pose space, SE(3), 
recursively expanding cells to obtain a finer pose discretization and discarding pose cells which can be guaranteed to not include the true pose.
The result is a discrete set of pose cells of which one of them encompasses the true pose.
This way, our method naturally detects ambiguities, irrespective of whether they arise from a lack of informative measurement points or object symmetries.
This ensures that non-experts are able to easily perform reliable fixture calibrations.

Additionally, we derive the likelihoods of poses within the pose superset, enabling confidence intervals which are tighter than the bounds.
The combination of bounds, confidence intervals and pose distributions makes our method relevant both in situations where a single fixture pose is needed with guaranteed or low-uncertainty accuracy, but also for potential future work where the expressive pose distributions open the possibility for downstream tasks to actively account for calibration uncertainties.

%The paper is structured as follows: First Section \ref{sec:relatedwork} presents related work on both contact- and vision-based pose estimation. Then, Section \ref{sec:methods} presents our method, followed by Section \ref{sec:experiments} which presents our evaluation producedure as well as the evalution results on both simulated and real experiments. In order to facilitate the use of our calibration method, Section \ref{} provides a brief guide to using our system. This is followed by \ref{sec:futurework} and \ref{sec:conclusion} which discusses limitations and future work, and concludes on the paper, respectively.

% \rlha{
% Reconfiguration of robot cell requires pose estimation of fixtures, etc.
% Manual pose labeling is time consuming and error prone.
% Vision-based solutions requires additional tools (camera) and setup, possibly markers. 
% Usually no guaranteed bounds.
% }
\section{Related Work}
\label{seq:relatedwork}
Fixture calibration can be done in many ways.
The two primary approaches are 
vision-based, where a vision system is installed and calibrated in the cell;
and contact-based, where the robot's internal encoders and forward kinematics are used to measure points on the fixture's surface.

Most contact-based methods assume that the user provides correspondences between measured surface points and features on the CAD model.
Some of the methods are tailored to specific sets of correspondences.
For example, \cite{thompson1958exact} provides a closed-form least squares solution, when exactly three point-to-point correspondences are available, 
and \cite{Zhang3points2008} proposes a calibration method, which also requires exactly three correspondences, but consisting of one point-to-point, one point-to-line and one point-to-plane correspondence. 

Other methods represent more general approaches.
\cite{horn1988closed} provides a closed-form least-squares solution for three or more point-to-point correspondences. 
\cite{olsson2006registration} extends this approach to allow point-to-line and point-to-plane correspondences using an iterative branch and bound algorithm to avoid getting stuck in local minima,
and \cite{Olsson2008Solving, Briales20174NonMinimalWithLagrangianDuality} improves on this approach using Lagrangian relaxation.

While the combination of the three correspondence-types do reflect a degree of generality and flexibility,
the methods still requires the user to provide accurate correspondences.
For objects consisting only of smoothly varying surfaces, none of the three correspondence types are suitable, and while the user might attempt to provide correspondences in such cases, they would be associated with significant and unknown uncertainties.
Even for objects, where the three correspondence types are suitable, there may still be a not insignificant amount of uncertainty related to the correspondences.
How accurately can a human mark a point-to-point or point-to-line correspondence?
Also, for near-symmetrical objects or objects with other ambiguities, the user might unknowingly provide correspondences which are entirely wrong.

If dense surface samples are available, local 3D features can be utilized to estimate the pose~\cite{chua1997point, johnson1999using, buch2017rotational} without requiring the user to annotate correspondences, however, moving a tool-tip across a surface provides a curve - not a surface, so methods which require dense samples are not directly applicable to contact-based pose estimation. 

Recent work~\cite{ykICRA2023} proposes and validates a framework for automatically finding point correspondences using point pair features \cite{DrostPPF2010} on sampled curves on the object surface.
The user does thus not have to manually annotate correspondences, and they show, that they are robust to correspondence outliers, however, they do not provide any estimate of the uncertainty of the point estimate they obtain.

To the best of our knowledge, our method is the first fully correspondence-free, contact-based pose estimation method, and the first to provide guaranteed bounds on SE(3).

%A likely reason why correspondences-free pose estimation has not previously been used with contact-based methods, is that correct, albeit inaccurate, point correspondences can be reliably established when using manual annotation. 
There are also vision-based approaches to fixture calibration. 
%Specifically, vision-based fixture calibration can be interpreted as an instance of visual object pose estimation.
If depth sensors are available, the aforementioned methods using dense surface samples can be used.
However, since the obtained point clouds also contain background points, it is not clear how any guarantees can be made for the resulting pose, even with guarantees for the depth errors.

Methods using color images typically either estimate correspondences as part of a pose estimation method \cite{surfemb, haugaard2023multi}, use strategies that compute similarities between the observed vision data and a vast set of hypotheses\cite{hinterstoisser2012model, ulrich2012combining, hagelskjaer2019using}, or directly regress object poses using deep learning~\cite{xiang2018posecnn, labbe2020cosypose, Wang_2021_GDRN}.

%This is rarely the case in the field of computer vision where images are acquired in uncontrolled environments and the use of manual annotations is rare. Therefore, most vision-based pose estimation methods make no assumption of known point correspondences and instead either explicitly estimate correspondences as a part of the pose estimation algorithm\cite{surfemb}, use strategies that compute similarities between the observed vision data and a vast set of hypotheses\cite{hinterstoisser2012model, ulrich2012combining, hagelskjaer2019using}, or directly regress object poses using deep learning\cite{xiang2018posecnn}.

For the methodologies described above, it is by far most common to provide a single best estimate of the pose, and while state of the art single pose estimation methods~\cite{surfemb, labbe2020cosypose, Wang_2021_GDRN} achieve impressive results on standard pose estimation benchmarks \cite{hodavn2020bop}, they are unable to express uncertainties arising from visual ambiguities such as those imposed by object symmetries.

Recently, there has been an increased interest in rotation and pose distribution estimation, defined as the estimation of a probability density function on SO(3) and SE(3), respectively. 
State of the art methods for both rotation \cite{murphy2021implicit, kipode2022iversen, hofer2023hyperposepdf} and full pose \cite{haugaard2023spyropose} distribution estimation provide a function for estimating the unnormalized likelihood of a pose hypothesis, and estimate the relative likelihoods for a large amount of samples on SO(3) or SE(3). 
This approach has resulted in highly expressive, non-parametric multi-modal distributions on SO(3) which with a hierarchical grid, proposed in \cite{murphy2021implicit, haugaard2023spyropose}, has been expanded to work on SE(3).
Because of the inherent complexity in natural images, however, to the best of our knowledge, no vision-based pose estimation method has provided guaranteed bounds on pose estimates.

A hierarchical grid on SE(3) is also at the core of our proposed method,
however, we use it in the setting of contact-based pose estimation, 
where it is reasonable to assume bounds on sampling errors, 
and we show, that we can propagate these bounds to pose space.

\newcommand{\posbound}{b_\text{p}}
\newcommand{\rotbound}{b_\text{r}}
\newcommand{\samplebound}{b_\text{s}}
\newcommand{\epsbound}{b_\epsilon}
\newcommand{\totalbound}{b}
\newcommand{\given}{\;|\;}
\newcommand{\frobot}{\text{robot}}
\newcommand{\fobj}{\text{mesh}}
\newcommand{\pose}{x}
\newcommand{\surface}{\mathcal{S}}
\newcommand{\poseset}{\mathcal{X}}
\newcommand{\dposeset}{\Tilde \poseset{}}
\newcommand{\translationset}{\mathcal{T}}
\newcommand{\recursion}{\tau}

\section{Methods}
\label{sec:methods}
Our method is based on hierarchical pose grids on \SET{}.
For each level in the grid, we can calculate bounds on the maximum point-to-mesh distance which can be caused by the discretization error in the grid.
Poses with at least one point-to-mesh distance which cannot be explained by discretization and sampling noise can then be discarded,
and only the remaining poses proceed to the next level in the hierarchical grid of higher pose resolution.

\subsection{Problem Formalization}
Let $P = \{p_1, p_2, \dots, p_n\}$ be a set of surface points relative to a robot's base frame, 
obtained by probing the surface of a fixture with a calibrated tool-tip attached to the end-effector of the robot.
Due to various uncertainties in the forward kinematics from the base of the robot to the tool-tip, the obtained points are not exactly on the surface of the fixture.
Let $p_i^*$ be the point on the fixture's surface which is closest to $p_i$:
$p_i^* = \argmin_{p \in \surface{}} ||p - p_i||$, where $\surface{}$ denotes the fixture surface.
The point-to-mesh error is then defined as $e_i = ||p_i - p_i^*||$.
Given a bound on the point-to-mesh error, $e_i \leq \totalbound{}\; \forall i$,
there exists a unique set of poses which satisfy the point-to-mesh constraints:
\begin{equation}
    \poseset{} = \left\{\pose{} \in \SET{} \given D(\surface{}, \pose{}, p_i) \leq \totalbound{} \quad \forall i \right\},
\label{eq:pose-set-def}
\end{equation}
where $D(\surface{}, \pose{}, p_i)$ is the point-to-mesh distance from the point $p_i$ to the fixture surface $\surface{}$ given the pose $\pose{}$. % = (R, t)$, $R \in \R^{3\times3}$, $t\in\R^{3\times 1}$.

We aim to find a tight superset $\hat \poseset{} \supset \poseset{}$, 
to be able to guarantee bounds on the true pose, $\pose{}^\dagger = (R^\dagger, t^\dagger)$.

\subsection{Initial Positional Bound}
\label{sec:position-aabb}
First, an initial bound on the fixture's position is found. 
Without loss of generality, we define the fixture's origin to be at the center of the the smallest sphere enclosing the fixture's mesh. 
We find this sphere from the fixture's 3D-model using the method described in \cite{fischer2003fast}. 
The radius of the sphere will be referred to as the fixture radius, $r$.

It follows from the point-to-mesh constraints, 
that the distance 
    from the sampled points 
    to the fixture position, $t^\dagger$,
is bounded by the sum of 
    the fixture radius, $r$,
    and the maximum sample error, $\samplebound{}$.
    
% First, an initial bound on the fixture's position, $t^\dagger \in \R^3$, is found.
% Let's define the center and radius of the fixture to be the center and radius of the smallest sphere enclosing the fixture's mesh.
% To find this sphere, we utilize the method described in \cite{fischer2003fast}.
% We then place the fixture's reference frame in the center of the fixture.

% It immediately follows from the point-to-mesh constraints, 
% that the distance 
%     from the sampled points 
%     to the fixture position, $t^\dagger$,
% is bounded by the sum of 
%     the fixture radius, $r$;
%     and the maximum sample error, $\samplebound{}$.
Intuitively, the set of fixture positions satisfying these constraints is the intersection of spheres with radius $r + \samplebound{}$ centered around each sampled point, $p_i$.
Formally, the set is defined by
\begin{equation}
    \translationset{} = \left\{ t\in \R^3 \given ||t - p_i|| \leq r + \samplebound{} \quad \forall i \right\}.
\end{equation}
Finding the Axis-Aligned Bounding Box (AABB) of $\translationset{}$ can be formalized as six Second-Order Cone Programs (SOCP):
\begin{equation}
    \begin{split}
        \underset{t}{\text{min}} & \quad c^\top t\\
        \text{subject to} & \quad ||t - p_i|| \leq r + \samplebound{} \; \quad \forall i,
    \end{split}
\end{equation}
with two programs per axis with varying $c$.
Specifically, the AABB of $\translationset{}$ is given by the AABB of the solutions to $c \in \{\hat i, -\hat i, \hat j, -\hat j, \hat k, -\hat k\}$,
where $\hat i, \hat j, \hat k$ are the orthonormal axis vectors.
We use the SOCP solver in CVXOPT~\cite{andersen2020cvxopt}.
The AABB provides an initial bound on the position, and the positional part of the hierarchical pose grid is initialized such that it encompasses the AABB.

\subsection{Discretization error bound on the $\R^{3}$ grid}
\label{sec:disc-err-pos-grid}
%A pose grid is the cartesian product between a positional grid in $\R^3$ and rotational grid in \SOT{}.
%
%and for the positional grid, we simply use a cubic octree.

The positional part of the hierarchical \SET{} grid 
is an octree where each node represents a cube in a regular grid.
For each recursion, each cube is divided into eight equal cubes.
We let each cube be represented by its center point,
and the largest distance from the center to any point in the cube is the distance to a cube corner, $\sqrt{3}l/2$, where $l$ is the side length of the cube.
Since $l$ is halved at each recursion, we can write the bounds of the positional discretization error as
$\posbound{} = \sqrt{3}l_0/2^{\recursion+1}$, where $l_0$ is the side length of cubes at recursion zero, and $\recursion$ is the recursion level.

\subsection{Discretization error bound on the \SOT{} grid}
For the rotational part of the grid, we use the HealPix grid \cite{gorski2005healpix} extended to \SOT{} by \cite{yershova2010generating}.
Specifically, HealPix defines a grid on the two-sphere,
which describes two of the three rotational degrees of freedom; azimuth and elevation.
\cite{yershova2010generating} proposes to take the cartesian product of the HealPix grid and a one-dimensional grid, describing the last rotational degree of freedom: Tilt.

HealPix provides the largest angular distance from the center of a rotation cell to any rotation in the cell, however, the angular distance incurred by tilt must also be taken into account.
Let $\theta$ be the largest angular distance from HealPix, 
and let $\phi$ be maximum tilt distance, which is simply half of the tilt resolution.
The maximum angle between a rotation cell center and any rotation in the cell is then given by
\begin{align}
\gamma
    &= 
    \max_{\alpha \in [0,\theta], \beta \in [0,\phi]}
    \cos^{-1}\left[\frac{\text{tr}\hspace{-0.1em}\left[R_x(\alpha)R_z(\beta)\right] - 1}{2}\right] \\
    &=
    \max_{\alpha \in [0,\theta], \beta \in [0,\phi]}
    \cos^{-1}\left[\frac{C_\beta + C_\alpha C_\beta + C_\alpha - 1}{2}\right] \\
    &=
    \cos^{-1}\left[\frac{C_{\beta^*} + C_{\alpha^*} C_{\beta^*} + C_{\alpha^*} - 1}{2}\right],
\label{eq:rot-disc}
\end{align}
where $C_\bullet = \cos \bullet$, $\bullet^* = \min(\bullet, \pi)$, and
\begin{equation}
    R_x(\alpha) R_z(\beta) =
    \begin{bmatrix}
        1 & 0 & 0\\
        0 & C_\alpha & -S_\alpha\\
        0 & S_\alpha & C_\alpha
    \end{bmatrix}
    \begin{bmatrix}
        C_\beta & -S_\beta & 0\\
        S_\beta & C_\beta & 0\\
        0 & 0 & 1
    \end{bmatrix}.
\end{equation}
Note that while we use $R_x$ here, it holds for rotations around any vector in the xy-plane.
\autoref{eq:rot-disc} provides a bound on the angular discretization error.

Recall the bound on the true fixture center, found in \autoref{sec:position-aabb}.
Let $\hat t$ denote the AABB center and let $b_t$ denote half of the AABB diagonal, which is the maximum distance from $\hat t$ to the true translation, $t^\dagger$.
It follows that the distance between the true fixture center, $t^\dagger$, and a given sample, $p_i$ is bounded: $||t^\dagger - p_i|| \leq ||\hat t - p_i|| + b_t$.
Also, the distance between a point on the unit circle before and after a rotation of $\gamma$ radians around the origin is $\sqrt{2 - 2\cos\gamma}$.
%\begin{align}
%    ||p_1 - p_0|| 
%        & = \sqrt{(\cos\theta - 1)^2 + (\sin\theta)^2} \\
%        & = \sqrt{\cos^2\theta + 1 - 2\cos\theta + \sin^2\theta} \\
%        & = \sqrt{2 - 2\cos\theta}
%\end{align}
The sample-wise bounds on the point-to-mesh errors related to rotation discretization are thus given by
\begin{equation}
    \rotbound^{i} = \left(||\hat t - p_i|| + b_t\right)\sqrt{2 - 2\cos\gamma}.
\end{equation}

\subsection{Distance constraints}
Given the bounds for the point-to-mesh errors related to 
discretization errors for the positional, $\posbound{}$, and rotational, $\rotbound{}$, part of the \SET{} grid,
as well as sampling, $\samplebound{}$, and numerical precision, $\epsbound{}$, 
we can provide total bounds on the point-to-mesh errors, as required in \autoref{eq:pose-set-def}:
\begin{equation}
\label{eq:total-bound}
    b^i = \posbound{} + \rotbound^i + \samplebound{} + \epsbound{}.    
\end{equation}
Given a pose hypothesis, 
the sampled surface points are transformed into the fixture frame, 
and the nearest distance to the mesh can be computed in sub-linear time using a tree to represent bounding volumes of mesh triangles.
We use Open3D's~\cite{open3d} implementation, built on Intel Embree~\cite{embree}.
If a pose cell has at least one point-to-mesh distance larger than the bound, the whole pose cell can be rejected.

\subsection{Expansion Strategy}
The pose superset $\hat \poseset{}$ is found by recursively expanding the SE(3) grid and rejecting cells using the criteria in \autoref{eq:pose-set-def}. 
The lowest recursion of the grid consists of the cartesian product of the positional grid at recursion 0 (a single cube containing the AABB) and the rotational grid at recursion 0 (the 72 rotations corresponding to the lowest recursion of the SO(3) HealPix grid).

Our expansion strategy utilizes that the rotational and positional parts of the grid do not have to be expanded simultaneously. Instead, only the part of the grid which incurs the largest point-to-mesh error bounds are expanded: The rotational grid is expanded if $\rotbound{} < \posbound{}$ and vice versa.

In normal cases, this expansion continues recursively until the sampling bound, $\samplebound{}$, becomes the dominating term in \autoref{eq:total-bound}. At this point very few of the expanded cells can be rejected causing the number of cells to increase exponentially.

% \rlha{
% Hvis du vil, T, kan du skrive det her ud:

% Vi starter med at have vores positions bound som en enkelt cubic cell omkring AABB.
% Vi har ingen rotationsbound, men rummet er i sig selv boundet, så vi bruger bare recursion 0 i SO(3) griddet.

% Vi udnytter så, at vi ikke behøver expandere griddet både for rotationer og positioner på én gang.
% Vi expanderer den del af griddet, som har den største bound. 
% Det vil sige expandér position, hvis $\rotbound < \posbound$, og ellers rotation.

% Vi observerer generelt, at søgningen kan fortsætte indtil $\samplebound$ er the dominating term in \autoref{eq:total-bound},
% hvorefter kun en meget begrænset del af ekspanderede celler kan rejectes, og antallet af celler stiger derfor eksponentielt. 

% I praksis tillader vi typisk 1e6-1e7 celler, og metoden tager fra et par sekunder til et minut på en bærbar pc med 16gb ram og 10 threads.
% }

\subsection{\SET{} Point Estimate with Guaranteed Bounds}
The found superset of poses which explain the measured points, $\hat \poseset{} \supset \poseset{}$, describe complex bounds on the true pose.
In case of multi-modal ambiguities, e.g. from a symmetric fixture, this ambiguity can be conveyed to the user, as will be discussed in \autoref{sec:exp-sim}.
However, when $\hat \poseset{}$ is uni-modal, it makes sense to provide a point estimate with guaranteed error bounds, to provide information which is more easily interpretable than the superset itself.

The positional point estimate is given by the center of the smallest sphere which encloses all possible positions in $\hat \poseset{}$. The positional error bound is given by the radius of the same sphere.
For rotation, we approximate the center of rotations by the center of the smallest enclosing sphere of quaternion representations of the rotations in $\hat \poseset{}$, and then calculate the maximum angular distance from the estimated center to all rotations in $\hat \poseset{}$.

\subsection{Estimation of Pose Likelihoods}
The previous sections involved the computation of guaranteed bounds on the pose. 
When only the point-to-mesh distance bounds are used, the resulting pose bounds do not necessarily improve significantly with more samples.
In consequence, large parts of the superset, $\hat \poseset{}$, while possible, may be highly unlikely given the surface measurements.
For this reason, we also consider a probability distribution over the superset of poses,
with the aim of determining confidence intervals which are tighter than the bounds.

The following derives the relationship between the surface point measurements, $P = \{p_1,p_2,\dots,p_n\}$, and the pose distribution $\p(\pose{} | P)$. 
The relationship between the surface measurements and the pose can be expressed using Bayes theorem: 
$\p(\pose{} | P) = \p(P|\pose{})\p(\pose{})\p(P)^{-1}$.
%
% Assuming that the surface point measurements are conditionally independent given $\theta$, this we can rewritten as:
%
% \begin{align}
%     \p(\theta | p_1,...,p_n) = \frac{\p(\theta)}{\p(p_1,...,p_n)}\prod_{i=1}^n \p(p_i|\theta)
% \end{align}
%
Assuming no prior knowledge of the pose, both $\p(\pose{})$ and $\p(P)$ are constants. 
It is also reasonable to assume that the surface point measurements are conditionally independent given the pose, $\pose{}$, which leads to the expression:
\begin{align}
\p(\pose{} | P) \propto \prod_{p_i\in P}\p(p_i|\pose{}).
\label{eq:proptoProd}
\end{align}
Using the law of total probability, the distribution corresponding to a single measurement can be factorized as
\begin{align}
    \p(p_i|\pose{}) = \int_{s_i \in \surface{}} \p(p_i|\pose{},s_i)\p(s_i|\pose{})ds_i,
    \label{eq:lawOfTotalProb}
\end{align}
where $\surface{}$ is the fixture surface and $s_i$ is the unknown true surface point for which $p_i$ was measured. 

If there is no prior knowledge of what the true surface point is, $\p(s_i|\pose{})$ can be treated as a uniform distribution over the fixture surface, and consequently as constant in the integral of \autoref{eq:lawOfTotalProb}. If we assume the measured points to be normally distributed, $p_i = s_i + \epsilon_i$, with $\epsilon_i \sim \N(0,\Sigma)$, it follows that $p_i$ and $\pose{}$ are conditionally independent given $s_i$: $\p(p_i|\pose{},s_i) = \p(p_i|s_i) = \N(p_i;s_i,\Sigma)$. Substituting this into \autoref{eq:lawOfTotalProb} yields
\begin{align}
    \p(p_i|\pose{}) = \int_{s_i \in \surface{}} \N(p_i;s_i,\Sigma)ds_i.
    \label{eq:intNormDist}
\end{align}

The integral in \autoref{eq:intNormDist} is expensive to compute. 
Fortunately, the measuring error of a robot tend to be much smaller than the size of the fixture which is to be calibrated. In this work, the measuring error is assumed to be isotropic, i.e. $\Sigma = \sigma^2 I$, so for most surface points $\N(p_i;s_i,\Sigma) \approx 0$. Since the major contribution to the value of the integral will be for the surface points closest to $p_i$, we choose to approximate the integral by the normal distribution corresponding to the closest point, $p_i^*$, given the pose, $\pose{}$. This leads to the following approximation of the likelihood:
\begin{align}
    \p(p_i|\pose{}) \approx \N(p_i;p_i^*,\sigma^2 I) \propto \exp\left(-\frac{\|p_i-p_i^*\|^2}{2\sigma^2}\right)
    \label{eq:norm}.
\end{align}
%The acute reader might notice that the approximation in Eq. \ref{eq:norm} will lead to an overestimation of likelihood at extremities, such as fixture corners, and an underestimation of likelihood in cavities. However, in general it is desirable to make surface measurements at extremities, so a user of the system should be asked to prioritise making such measurements. In this case the use of a uniform prior $\p(s_i|\pose{})$ will leads to an underestimation of likelihood at extremities and vice versa with cavities. Consequently the assumption of a uniform prior and the approximation in Eq. \ref{eq:norm} has opposite effects that to some extend cancel each other out.
Since $p_i^*$ can be found efficiently (see \autoref{sec:disc-err-pos-grid}),
substituting \autoref{eq:norm} into \autoref{eq:proptoProd} provides a relative pose likelihood function which is efficient to compute:
\begin{align}
    \p(\pose{} | P) & \propto \prod_{p_i\in P} \exp\left(-\frac{\|p_i-p_i^*\|^2}{2\sigma^2}\right).
    \label{eq:rel-pose-likelihood}
\end{align}

\subsection{Confidence Intervals}
We obtain a discrete pose distribution by stratified Monte Carlo samples, sampling $k$ poses uniformly within each cell of $\hat \poseset{}$.
Specifically, the relative probabilities are estimated using \autoref{eq:rel-pose-likelihood}
and normalized over the samples.
The result is a discrete set of poses, $x_i \in \dposeset{}$, with associated, estimated probabilities, $\p(x_i|P)$. 

The availability of such a distribution motives a new point estimate of the pose, 
namely the expected pose.
We compute the expected position and rotation as 
\begin{equation}
    \hat t = \sum_{\pose_i \in \dposeset{}} \p(\pose_i|P) t_i
    \quad \text{and} \quad
    \hat q = \sum_{\pose_i \in \dposeset{}} \p(\pose_i|P) q_i,
\label{eq:expected-pose}
\end{equation}
respectively, where $t_i$ is the translation of pose $\pose_i$,
$q_i$ is a quaternion representation of the rotation in $\pose_i$,
and all $q_i$'s are mapped to the same quaternion hemisphere. 
This would cause problems with very large rotational ambiguities such as in case of symmetries, however, then the confidence intervals are very large anyway.
Normalization of $\hat q$ is left out for clarity.
Confidence intervals are then found as the error for which the desired amount of probability mass is within that error from the expected pose.
% Simulation
%   [x] Bounds, symmetry figure
%   [x] Conf. intervals
%   [x] pose bound vs sample bound plot
% Real
%   [ ] Bounds and confidence intervals (MATLAB, expert vs non-experts)
\section{Experiments}
\label{sec:experiments}

We evaluate our method both in simulation and on the robot shown in \autoref{fig:system}.
To evaluate the ability to represent potential ambiguities,
we run experiments on objects of the SYMSOL~\cite{murphy2021implicit} dataset with varying degrees of symmetry.
We also evaluate bounds and confidence intervals in simulation, where the true pose is known.
On the real system, one expert as well as three non-experts, to demonstrate ease of use, measure points on the MATLAB logo shown in \autoref{fig:system}.

All experiments are run on a laptop with an Intel i7-8650U 1.90GHz CPU and 16GB RAM.
We allow up to $10^7$ cells when expanding the pose grid, and each experiment takes between a few seconds and two minutes, depending on geometry and how informative the sampled points are.

\subsection{Simulation}
\label{sec:exp-sim}

\begin{figure*}
    \centering
    \hfill
    \includegraphics[height=0.14\linewidth,trim={0 0 0 .5cm},clip]{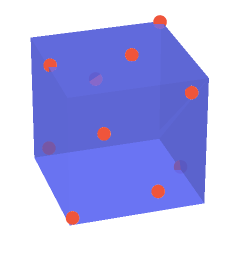}\hspace{1em}\hfill\hfill
    \includegraphics[height=0.145\linewidth,trim={0 .5cm 0 .5cm},clip]{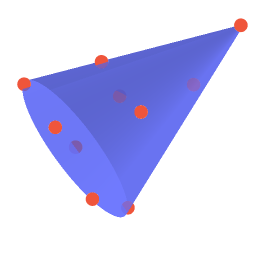}\hfill\hfill
    \includegraphics[height=0.14\linewidth,trim={0 0 0 .5cm},clip]{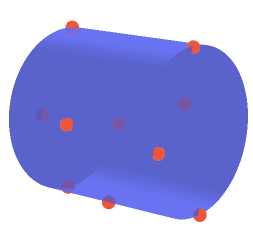}\hfill~\\
    \includegraphics[width=0.32\linewidth,trim={3cm 10.7cm 2.5cm 2cm},clip]{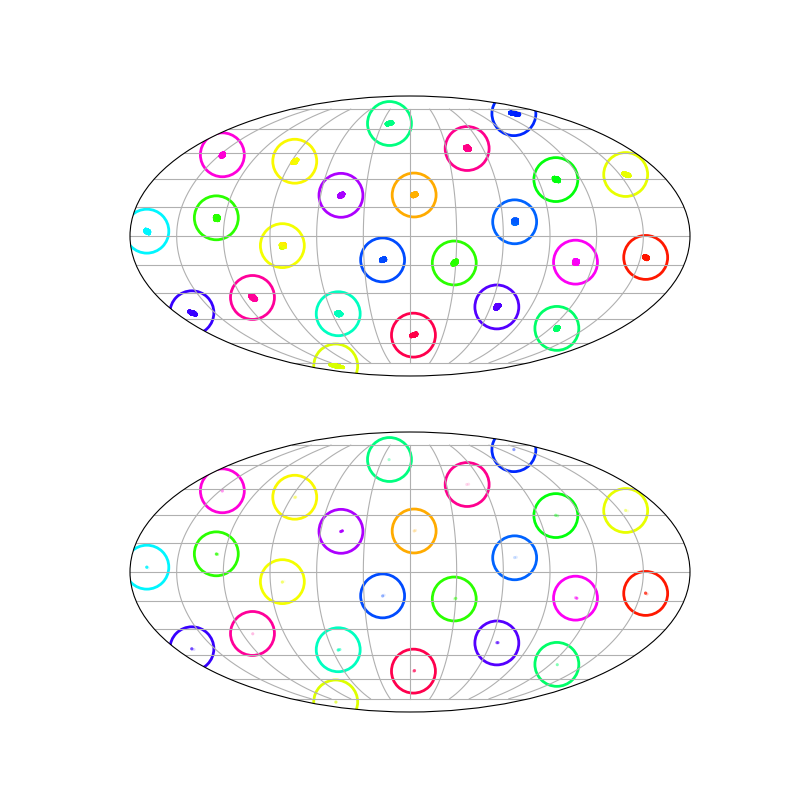}\hfill
    \includegraphics[width=0.32\linewidth,trim={3cm 10.7cm 2.5cm 2cm},clip]{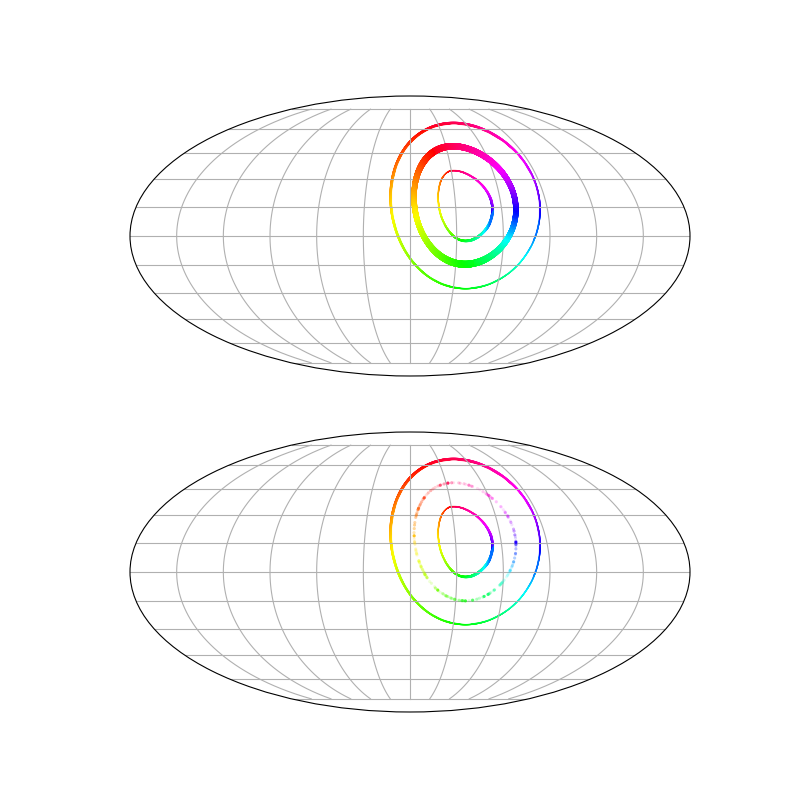}\hfill
    \includegraphics[width=0.32\linewidth,trim={3cm 10.7cm 2.5cm 2cm},clip]{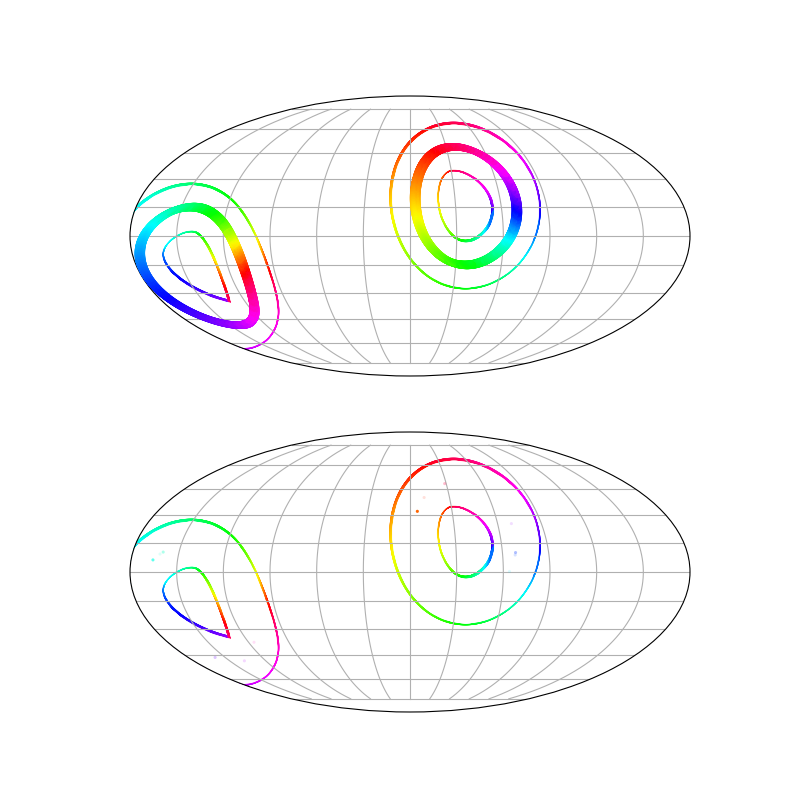}
    \caption{
        Qualitative pose estimation results of three different objects from the SYMSOL~\cite{murphy2021implicit} dataset.
        From left to right: A cube, a cone and a cylinder.
        Top: The objects are rendered semi-transparently in blue together with ten simulated surface samples with noise, $\samplebound = 0.3\text{mm}$, in red.
        From the CAD model and noisy surface samples, our method provides a set of poses which are \textit{guaranteed} to include the true pose.
        Bottom: We show the rotation part of the pose set with two rotational dimensions shown by the position on a sphere, shown in a Mollweide plot, and the last dimension visualized by color.
        The true rotation up to symmetry is shown by circles for discrete symmetries (cube) and "stroked" circles for continuous symmetries (cone and cylinder). 
        Our method naturally captures all the modes without any prior knowledge about symmetries, and this ambiguity can be conveyed to a user.
    }
    \label{fig:symsol}
\end{figure*}

To obtain simulation results, we simulate the manual surface sampling procedure.
First, we sample a large number of uniform samples from the object surface.
Second, we reduce the samples down to a small number, using farthest point sampling.
In our experiments, we reduce 1.000 uniform samples to 10 surface samples.
Finally, based on the sampling error bound, $\samplebound{}$, we add noise, $e$, to the samples.
Specifically, $e$ is sampled from a normal distribution, which is truncated at $||e|| = \samplebound{}$, 
and with covariance $(0.3 b_s)^2 I$.
For the simulation experiments, we normalize the meshes to have an enclosing sphere with a diameter of 25 cm to mimic typical fixture sizes.
Examples of simulated, sampled points are shown in \autoref{fig:symsol}.

To present our method's ability to provide multi-modal supersets, $\hat \poseset{}$,
we apply our method on objects from the SYMSOL~\cite{murphy2021implicit} dataset with varying degrees of symmetry.
The rotational part of the supersets is shown in \autoref{fig:symsol}.

\begin{figure}
    \centering
    \includegraphics[width=\linewidth]{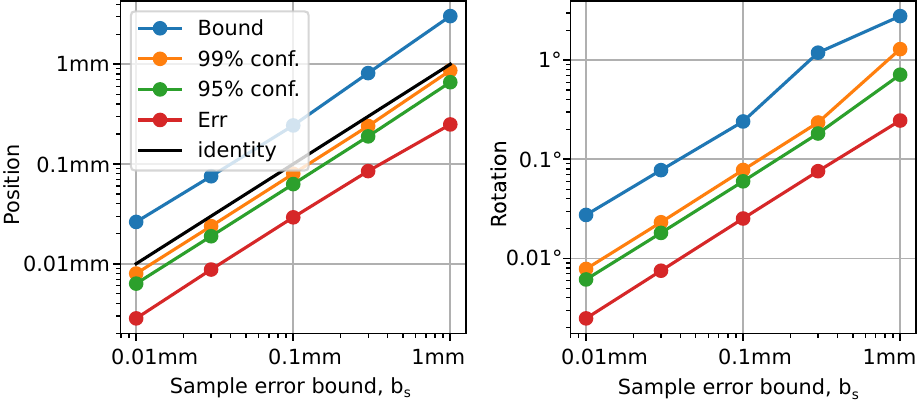}
    %{\footnotesize Sample error bound, $\samplebound{}$ [m]}
    \caption{
        Pose uncertainty for simulations on the MATLAB logo with varying sample error bounds.
        The true error for the expected pose (\autoref{eq:expected-pose}) is shown in red and is within the confidence intervals and bounds.
        The pose uncertainties decrease proportionally with the sample error bound.
    }
    \label{fig:meas-err-plot}
\end{figure}

When the superset is uni-modal, the bounds and confidence intervals which can be obtained, become of interest.
To this end, we simulate experiments on the MATLAB logo, shown in \autoref{fig:system},
with varying sample error bounds.
The results are shown in \autoref{fig:meas-err-plot}.
The positional errors of the expected poses are lower than the sample error bounds and well below both the confidence intervals and pose bounds.
We observe similar results for other fixtures.

\subsection{Real}
For the experiments on the real system,
we first propose a tool to measure points and provide methods for calibrating the tool tip as well as for estimating the sample error bound.
Then we run experiments with one expert and three non-experts to demonstrate ease of use of the proposed method.

To make sure there is contact between tool tip and surface at the time of point sample collection,
samples are collected by positioning the tool tip close to a surface, and letting the robot tap the surface with a force-based controller.

\subsubsection{Tool design}
\begin{figure}
    \centering
    \includegraphics[width=0.5\linewidth]{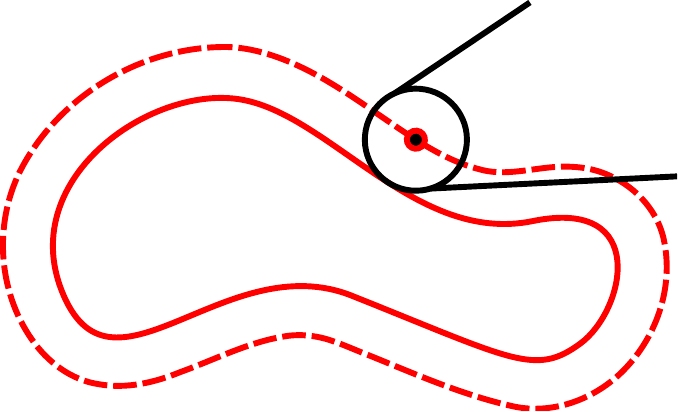}
    \caption{
        The tool-tip with the steel ball is shown in black.
        The fixture is shown in red, and the Minkowski sum of the steel ball and the fixture is illustrated by a dashed, red line.
        The center of the steel ball lies on the Minkowski sum surface independently of the angle of sampling.
    }
    \label{fig:minkowski}
\end{figure}

We design a low-cost point measuring tool which can be seen in \autoref{fig:system} and \ref{fig:minkowski}.
The tool consists of an 3D-printed cone and an off-the-shelf 3 mm steel ball.
The cone is attached to the robot's end effector,
and an indent at the tip of the cone allows the steel ball to be glued firmly onto the cone.
This way, we get the flexibility of 3D-printing,
for easy incorporation in tool designs, 
and the tolerances and wear-resistance of a steel ball.

\subsubsection{Tip calibration}
We first obtain a coarse calibration by approximately rotating the tip around a fixed point with respect to the robot's base frame and record robot configurations to solve the set of linear equations with least squares:
\begin{equation}
     \left\{T_{\base}^{\tcp}\right\}_i p_{\tcp}^{\tip} = p_{\base}^{\tip}
      \quad \forall i,
\end{equation}
where $p_{\tcp}^{\tip}$ and $p_{\base}^{\tip}$ are constant but unknown.
A coarse table normal calibration can then be obtained by fitting a plane to three non-colinear table points.

From the coarse tip- and table calibrations, the robot can automatically gather a larger dataset.
To get an accurate estimate of the table's normal, a set of points on the table is collected with a fixed tcp orientation.
Specifically, we subtract the centroid from the table points, apply SVD and use the vector corresponding to the least singular value to obtain $\hat n_{\base}$,
which is independent of the coarse calibration.%We thus don't assume the coarse calibrations to be accurate.

Then, to obtain a better calibration of the tip, we obtain a new dataset with varying orientations, providing the following equations, also solved with least squares:
\begin{equation}
    \hat n_{\base} \left\{T_{\base}^{\tcp}\right\}_i p_{\tcp}^{\tip} = k
     \quad \forall i, 
\label{eq:tip-calib-fine}
\end{equation}
where $p_{\tcp}^{\tip}$ and $k$ are constant and unknown.

This calibration provides the center of the steel ball, and it turns out that this is desirable.
When sampling points on the fixture surface, it is not clear which point on the steel ball is in contact with the surface, however, it is clear, that the center of the steel ball lies on the Minkowski sum~\cite{hadwiger1950minkowskische} of the fixture and the steel ball. See \autoref{fig:minkowski}. 
Since the steel ball and fixture geometries are known,
we can compute this Minkowski sum. We use the CGAL implementation~\cite{cgal:minkowski}.
We set the sample error bound above the largest residual observed in \autoref{eq:tip-calib-fine}: $\samplebound{} = 1.0~\text{mm}$. %, which is also the global accuracy according to the robot's datasheet

\subsubsection{User Study}

We gave three students, which were unfamiliar with our method, a 1-minute explanation of the task and asked them to sample 15 surface points on the setup shown in \autoref{fig:system}.
The resulting positional bounds were 3.5, 2.7 and 3.1~mm,
and the rotational bounds were 2.7\textdegree{}, 1.6\textdegree{} and 3.9\textdegree{}.
The 99\% confidence intervals were 0.80, 0.81 and 0.83~mm for position and 0.73\textdegree{}, 0.50\textdegree{} and 0.99\textdegree{} for rotation.
The expected poses were all within 0.87 mm and 0.74\textdegree{} of each other.
The expert obtained bounds of 1.6 mm and 1.2\textdegree,
and 99\% confidence intervals of 0.39 mm and 0.26\textdegree.

%Make simple, short video instruction and include in supplementary material.
%For reproducibility, this should be the only communication / training to the users.
%Include a text-version in the paper to show how little instruction it is.

%\subsection{Pose Estimation}
%For all users, for all objects, (with and without guidance) perform the pose estimation and record
%time to sample points, method runtime, bounds.

\section{Limitations and Future Work}
\label{sec:futurework}
One limitation of our method is the assumption of guaranteed bounds on the sample error. The sample error can be caused by everything from flex in the robot or fixture to bias in the robot's internal encoder - and setting the sample error bound too conservatively will lead to large pose bounds.

A current limitation is that the user may not know what points to sample to reduce the bounds.
The user would obtain this intuition from trial and error, but
identifying points which are informative, and showing them to the user, to build the intuition faster, would be an interesting future work.
%Note that this is \textit{not} to obtain correspondences, but simply to guide the user's intuition for which points to sample to obtain accurate calibrations.

This work considered calibration of a single fixture. 
However, the superset could  be computed for several fixtures given the same samples, and since an empty pose set would indicate mesh-samples incompatibility, our method could be used for classification as well as calibration.

Finally, our method is applicable for any sensor where guaranteed bounds on surface measurements can be provided, potentially making our method relevant for use with vision sensors in certain cases.

% \rlha{
% If the correspondences are not informative enough, the number of hypothesis increases exponentially.

% How to define guaranteed sample error.

% If guaranteed bounds can be provided for sensors, 
% the method can also be applied there (e.g. vision).

% Guided point sampling: automatically training users to pick more informative points.
% Note the difference between correspondences and guided sampling.
% Extend method to guided pose estimation with approximate correspondences which can resolve symmetries and help the user select good points.

% Local accuracy is much higher than global accuracy.
% If a lower, local bound could be determined, this could probably be used to improve pose bounds.

% If the CAD model is not the model from which the samples have been taken,
% the method would realize this, if no poses explain the sampled points.
% The method can thus also be used for object classification, and provide guarantees in case the sampled points can only be explained by one of more objects.
% }
\section{Conclusion}
\label{sec:conclusion}
This paper presented a novel, correspondence-free method for contact-based fixture pose calibration. 
The user simply has to measure a few points on the surface of the fixture, avoiding manual and error-prone correspondence annotations. 
The output of the method is a tight superset of the poses which could explain the measured points. 
%The superset naturally represents ambiguities caused by object symmetries and uninformative points. 
The superset provides \textit{guaranteed} bounds on the true pose and naturally represents ambiguities caused by symmetries and uninformative points.
%The computation of the bounds is made tractable by the use of a hierarchical grid on SE(3). 
Our method was evaluated both in simulation and on a real, collaborative robot, showing great potential for easier and less error-prone fixture calibration.

%\input{tex/todo.tex}

%\addtolength{\textheight}{-12cm}   % This command serves to balance the column lengths
                                  % on the last page of the document manually. It shortens
                                  % the textheight of the last page by a suitable amount.
                                  % This command does not take effect until the next page
                                  % so it should come on the page before the last. Make
                                  % sure that you do not shorten the textheight too much.

%\begin{table}[h]
% \caption{An Example of a Table}
% \label{table_example}
% \begin{center}
% \begin{tabular}{|c||c|}
% \hline
% One & Two\\
% \hline
% Three & Four\\
% \hline
% \end{tabular}
% \end{center}
% \end{table}

%   \begin{figure}[thpb]
%       \centering
%       \framebox{\parbox{3in}{We suggest that you use a text box to insert a graphic (which is ideally a 300 dpi TIFF or EPS file, with all fonts embedded) because, in an document, this method is somewhat more stable than directly inserting a picture.
% }}
%       %\includegraphics[scale=1.0]{figurefile}
%       \caption{Inductance of oscillation winding on amorphous
%       magnetic core versus DC bias magnetic field}
%       \label{figurelabel}
%   \end{figure}

{\small
\bibliographystyle{IEEEtran}
\bibliography{references}
}

\end{document}